\documentclass[reprint,amsfonts, amssymb, amsmath,showkeys,pra,nofootinbib, twocolum,longbibliography,aps, superscriptaddress]{revtex4-2}

\usepackage[dvipsnames,table]{xcolor}
\usepackage[caption=false]{subfig}
\usepackage{graphicx}
\usepackage{float}
\usepackage{tabularx}
\usepackage{multirow}
\usepackage{comment}
\usepackage{braket}
\usepackage[colorlinks=true,
            linkcolor=blue,
            citecolor=blue,
            urlcolor=blue]{hyperref}

\newcommand{\bsc}{Barcelona Supercomputing Center (BSC)}
\newcommand{\upf}{BCN Medtech, Universitat Pompeu Fabra, Barcelona, Spain}
\newcommand{\uba}{Departament de Física Quàntica i Astrofísica, Facultat de Física,
Universitat de Barcelona, E-08028 Barcelona, Spain}
\newcommand{\ubb}{Institut de Ciències del Cosmos, Universitat de Barcelona,
ICCUB, Martí i Franquès 1, E-08028 Barcelona, Spain.}
\newcommand{\icrea}{ICREA, Barcelona, Spain.}

\begin{document}

\title{Towards interpretable AI with quantum annealing feature selection}

\author{Francesco Aldo Venturelli}
\affiliation{\upf}
\affiliation{\bsc}
\author{Emanuele Costa}
\affiliation{\bsc}
\author{Sikha O K}
\affiliation{\upf}
\author{Bruno Juliá-Díaz}
\affiliation{\uba}
\affiliation{\ubb}
\author{Miguel A. González Ballester}
\affiliation{\upf}
\affiliation{\bsc}
\affiliation{\icrea}
\author{Alba Cervera-Lierta}
\affiliation{\bsc}

\begin{abstract}

Deep learning models are used in critical applications, in which mistakes can have serious consequences. Therefore, it is crucial to understand how and why models generate predictions. This understanding provides useful information to check whether the model is learning the right patterns, detect biases in the data, improve model design, and build systems that can be trusted. This work proposes a new method for interpreting Convolutional Neural Networks in image classification tasks. The approach works by selecting the most representative feature maps that contribute to each prediction. To solve this combinatorial problem, we encode it into a quantum constrained optimization problem and propose to solve it using quantum annealing. 
We evaluate our method against the state-of-the-art explainable AI techniques, specifically GradCAM and GradCAM++,
and observe an improved class disentanglement, i.e. the model's decision boundaries become more distinct and its reasoning more transparent. This demonstrates that our approach enhances the quality of explanations, making it easier to understand which features the model relies on for specific predictions. In addition, we study the computational behavior of the quantum annealing algorithm. Specifically, we analyze the minimum energy gap of the system during computation and the probability that the algorithm finds the correct solution. These analyses provide theoretical insight into why the method works effectively in practice.

\end{abstract}

\maketitle

\section{Introduction}\label{sec:intro}

Deep learning (DL) has emerged as the standard approach to solving computer vision problems due to its ability to efficiently process large volumes of unstructured data such as images and videos~\cite{trigka2025comprehensive, noor2024survey}. One of the most relevant contributions of DL to the development of the area of computer vision is attributed to the introduction of the Convolutional Neural Network (CNN)~\cite{lecun1998convolutional,krizhevsky2012imagenet}. CNN sequentially applies convolution operations that learn local spatial correlations within data by reducing the data-dimension across multiple layers while extracting higher-level features.

Different CNN architectures have been proposed and explored so far. The first breakthrough in CNNs began with AlexNet, which achieved significant performance on the ImageNet~\cite{russakovsky2015imagenet} challenge and sparked a wave of architectural innovations~\cite{krizhevsky2012imagenet}. Subsequent architectures such as VGGNet~\cite{simonyan2014very} and ResNet~\cite{he2016deep} demonstrated that deeper structures and residual connections could significantly improve representation learning and training stability, becoming widely used backbone models for visual recognition systems~\cite{khan2020survey, rangel2024survey}. 
Subsequently, many other variants of CNNs have been proposed for various pixel-level tasks such as U-Net~\cite{ronneberger2015u} that introduced an encoder–decoder scheme with skip connections and became a standard approach for image segmentation~\cite{rayed2024deep, azad2024advances}. Modern CNNs are also used in generative modeling, including variational autoencoders and generative adversarial networks~\cite{kingma2015adam, goodfellow2014gan, han2019image} and diffusion models~\cite{sohl2015deep, ho2020denoising, rombach2022high, croitoru2023diffusion}.

Despite their success, DL models often behave as ``black boxes". Their predictions emerge from complex sequences of operations involving millions of trainable parameters, rendering internal decision processes difficult to interpret~\cite{zhang2018interpretable, buhrmester2021analysis, qamar2023understanding}. This lack of transparency poses a significant challenge in safety-critical applications where understanding model behavior is essential for establishing trust and accountability~\cite{doshi2017towards}. To address this limitation, the field of explainable AI (XAI) has emerged, offering a suite of techniques designed to inspect model outcomes and highlight the reasoning behind specific predictions~\cite{vonder2023analysis,salih2025perspective}.
XAI methods are broadly categorized into \textit{ante-hoc}, which incorporate interpretability constraints during model design, and \textit{post-hoc} approaches, which analyze trained models after the fact. 

In the field of interpretability, CNNs are relatively well-studied architectures, with a broad body of literature and a wide range of post-hoc techniques developed specifically for their analysis.
Among them, gradient-based activation mapping methods have gained considerable attention. Examples are Grad-CAM~\cite{selvaraju2017gradcam} and Grad-CAM++\cite{chattopadhay2018gradcampp}, which compute the gradients of a target class score with respect to the final convolutional layer, producing coarse localization maps that highlight image regions that most influence the prediction. 
These methods, however, operate at an aggregate level. They combine contributions from all feature maps (FMs) into a single saliency visualization without disentangling the role of individual FMs within the network. This aggregation obscures which specific learned representation is responsible for a given prediction.

A promising direction to address the limitations of these gradient-based mapping methods lies in feature selection (FS), a well-established paradigm that has gained renewed relevance in the context of XAI. Traditionally, FS has been predominantly employed as a data pre-processing step, used to reduce input dimensionality, remove noise, and improve training efficiency before model learning begins ~\cite{jovic2015review}. This conventional view has led to FS being underutilized as a tool for model interpretation. However, when applied to learned representations rather than raw input data, FS offers a powerful mechanism for post-hoc explainability that remains largely underexplored. Some recent works have recognized the potential of FS principles within the XAI paradigm \cite{ribeiro2016should,lundberg2017unified, vaswani2017attention, chen2018learning,yoon2018invase}.
However, it is known that FS is computationally challenging, since searching a subset of features that satisfies a certain condition scales exponentially with the number of features in the search space. For this reason, proposed approaches to address this problem are of a heuristic nature.

In this work, we explore the novel use of quantum computing for FS in the context of XAI. Quantum computing provides a natural framework for encoding combinatorial optimization problems, such as FS. In particular, by mapping the FS task into a Quadratic Unconstrained Binary Optimization (QUBO) formulation, each binary variable can represent the inclusion or exclusion of a feature, and the objective function can be designed to capture relevance, redundancy, and task-specific constraints.
Within this framework, quantum computers enable the efficient exploration of the exponentially large solution space. This process can be interpreted as a guided search over the space of feature subsets, where multiple candidate solutions are implicitly considered in parallel and progressively driven toward optimal or near-optimal configurations.

Previous approaches to reformulating FS schemes as QUBO problems have primarily focused on input-level features, largely overlooking the internal learned representations of deep networks~\cite{ferrari2022towards, mucke2023feature, bhagawati2023approach, hellstern2024quantum, romero2025quantum}. This gap motivates the exploration of quantum computing devices available in the current NISQ era \cite{bharti2022noisy}, which encompasses both digital and analog paradigms, including quantum annealers (QA)~\cite{Quinton2025} specially designed to solve combinatorial optimization problems, offering a promising avenue for tackling high-dimensional FS problems directly on learned representations, even in the absence of a proven quantum advantage. 

In this work, we propose a QA-based FS approach for generating the bit-strings containing the selected FMs:  the explanation maps of CNNs. By formulating FS as a QUBO problem, we leverage QA to efficiently identify the subset of FMs that most strongly contribute to a given prediction. The novelty we introduce relies in the fact that we do not apply FS directly on input pixels. Rather, we consider the hidden representations that the model learns at a particular stage as the elements over which performing the selection. Practically speaking, we enforce the QA to sample the bit-string that maximizes the linear term representing the importance of a single FM to the prediction, while minimizing the geometrical similarity among distinct FMs through the cosine similarity term, in a way similar to the well-known NP hard Maximum Diversity problem \cite{kuo1993analyzing, ghosh1996computational, marti2013heuristics}.
In this way, we combine the core aspect of GradCAM (gradient importance) with a proper quadratic term (that considers the mutual geometrical similarity) to construct the QUBO problem.

Reducing the number of FMs by FS schemes provides a natural strategy to mitigate scalability and dimensionality challenges in DL models. Additionally, it improves model interpretation by isolating FMs that strongly contribute to a given prediction, compressing and discarding similar and redundant representations. In this way we characterize the most informative patterns that the model learns during the training process to be able to interpret the resulting predictions. 

The article is organized as follows: in Sec. \ref{sec:results} we give a detailed description of the FS algorithm and define the mathematical formulations and the QA simulation procedure. Subsequently, we present the main results, i.e. the explanation maps sampled by QA and the class-class correlation maps that quantify the disentanglement among distinct image classes. Then, we analyze the model's success by defining physics-derived metrics, such as the minimum energy gap between the first excited state and the ground state of the QUBO Hamiltonian and we explain the meaning behind these results.
In sec. \ref{sec:discussion}, we discuss the relevance of our approach by explaining the key ideas behind the application of QA to the framework of interpretability, while highlighting our contributions, the novelty with respect to the state of the art, and the possible future directions that can be further explored. More details on the methods employed are presented in Sec.~\ref{sec:methods} and in the Appendices.

\section{Results}\label{sec:results}

\subsection{Quantum annealing feature map selection algorithm}

\begin{figure*}[t!]
    \centering
    \includegraphics[width=1\textwidth]{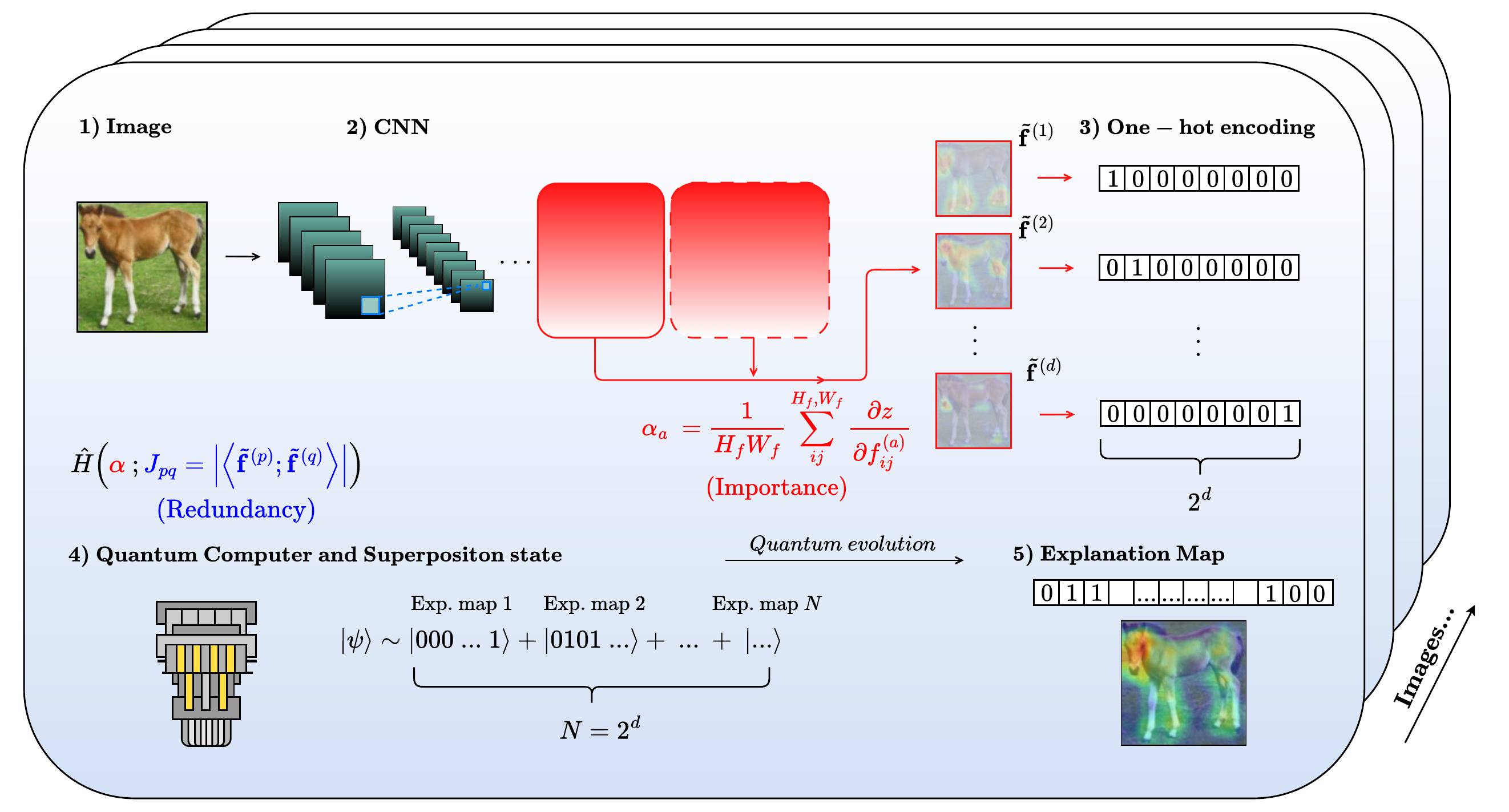}
        \caption{
        Representation of the FS algorithm reformulated as a QUBO problem. $1-2)$ A single image is fed into a CNN that has been previously trained. From the last convolutional layer we extract the $N_{f}$ FMs, $f_{ij}(x)$ and compute its gradient with respect to the cost function $z$. We call this quantity the \emph{importance} $\alpha_{a}$ for each $a\in\{1,\cdots,N_{f}\}$ FMs and keep only those with $\alpha_{a}>0$.
        3) A one-hot encoding is applied to the $d$ FMs that contribute positively to the gradient.
        4) We construct the QUBO Hamiltonian, that is formed by a linear term proportional to the importance and an interacting term, $J_{pq}$ that computes the cosine similarity between all pairs of the FMs, a measure of redundancy. We then prepare the initial quantum state in superposition of all possible $d$ combinations of the positive contributing FMs and we use QA to evolve the QUBO Hamiltonian towards its ground state.
        5) Finally, we sample the resulting bitstring, which corresponds to the explanation map containing the FMs that satisfy the constraints imposed by the problem Hamiltonian: maximizing the positive contribution of each FM while minimizing their geometrical similarity.
        This pipeline is repeated for the $M$ images from the test set.}
    \label{fig:MainFigure}
\end{figure*}

We propose a XAI method that identifies the relevant FMs needed to classify a given image. To do so, we encode the optimization problem of finding this set into a QUBO problem and solve it with QA.
The overview of our algorithm, as shown in Fig.~\ref{fig:MainFigure}, is the following: we take the FMs produced from the last convolutional block and select those whose gradient positively contributes to the model's output (importance measure). We then generate an Ising-type Hamiltonian using the absolute value of the cosine similarity between the FMs and the importance of each FM. The QA evolves the corresponding Hamiltonian to find the ground state containing the superposition of the FMs relevant for the image. By repeating this process for the whole image dataset, we extract a class-class correlation map that allows us to interpret which classes share common FMs. In other words, we disentangle the FMs from the CNN to obtain only those relevant to identify the image class.

Formally, a CNN is a parametric function composed of convolutional operators that map an image $\mathbf{x}\in \mathbb{R}^{H\times W\times C}$, where $H$ and $W$ are the height and width of the image (e.g. the number of pixels that it contains), and $C$ is the number of channels (e.g. $C=3$ for RGB), to hierarchical FMs, followed by a dense classifier operating on the learned representations. Each convolutional layer takes as input the FMs produced by the previous layer and applies learnable filters to extract higher-level features; the spatial dimensionality is reduced through sequences of striding or pooling operations. 

Given the set of FMs from the last convolutional layer, $\{\mathbf{f}\}\in \mathbb{R}^{H_{f}\times W_{f}\times N_{f}}$, where $H_{f}$ and $W_{f}$ are the height and width of the last layer (typically, much smaller than the image $H$ and $W$) and $N_{f}$ is the number of the learned representations (that depends on the CNN architecture, but typically much larger than $C$), and the final output of the dense classifier, $z=\text{CNN}(\mathbf{x})$, where $\text{CNN}$ represents the full CNN processing layers, we compute the following quantity
\begin{equation}
    \alpha_a=\frac{1}{H_{f} W_{f}}\sum_{ij}^{H_{f}, W_{f}} \frac{\partial z}{\partial f^{(a)}_{ij}},
    \label{eq:alpha}
\end{equation}
for each FM $a$, which is the global average pooling of the gradient. This quantity is equivalent to obtaining a scalar measure of the importance for each FM. In the GradCAM approach, $\alpha_a$ is just the weight of the corresponding FM. Depending on the sign of $\alpha_a$, we can distinguish class-confirming FMs (positive sign), which contribute to characterizing the class, and class-opposing ones (negative sign) that pivot on characterizations that lower the confidence of the classification. In our approach, we propose a \textit{filtered} GradCAM, i.e. $f$GradCAM, by selecting the learned representations that satisfy
\begin{equation}
    \{\tilde{\mathbf{f}}\} \equiv \{\mathbf{f}^{(a)}|\alpha_{a}>0\}.
\end{equation}
Notice that each $\tilde{\mathbf{f}}\in\mathbb{R}^{H_{f}\times W_{f}}$ and there will be $d\leq N_{f}$ filtered FM. 

Next, we compute the cosine similarity between the pairs of filtered FM, 
\begin{equation}
J_{pq}=| \langle \tilde{\mathbf{f}}^{(p)}; \tilde{\mathbf{f}}^{(q)}\rangle |=\frac{|\sum_{ij} \tilde{f}_{ij}^{(p)} \tilde{f}_{ij}^{(q)}|}{||\tilde{\mathbf{f}}^{(p)}||\  ||\tilde{\mathbf{f}}^{(q)}||}.
\label{eq:cos_sim}
\end{equation}
This quantity is large when there is overlap and, therefore, redundancy between distinct FMs, while it is small when the FMs are more geometrically independent. The idea behind the cosine similarity is to assume that FMs that encode similar information, which is redundant for the model, are parallel vectors in the feature latent space. Thus, we look for the subset of vectors that are maximally orthogonal to each other.

Our goal is to select the subset of positive representations maximizing the importance of the filtered FM, $\tilde{\alpha}>0$ (gradient) and minimizing the redundancy $J_{pq}$ (cosine similarity) between the filtered FMs. 

To do so, we opt for using QA due to its adaptability to solve combinatorial optimization problems, such as the FS problem. Initially, we need to convert the classical problem in its QUBO form to operate directly with the QA. Thus, once we collect the FMs extracted from the target convolutional block, we realize a one-hot encoding of the $d$ FMs that positively contribute to the gradient of the CNN's output, i.e. those with $\alpha_{a}>0$. We use a one-hot encoding for the filtered FM, representing each of them as a unary vector $\hat{n}_{p}\equiv\frac{1-\sigma_{z}^{(p)}}{2}$, where $\sigma_{z}=\mathrm{diag}(1,-1)$ is the Pauli-Z operation acting on qubit $p$. Therefore, we will need $d$ qubits to represent the space of filtered FMs. Next, we construct a QUBO Hamiltonian of the following form:
\begin{equation}
    \hat{H}_{\text{QUBO}} = (1-\beta) \frac{1}{2}\sum_{pq} J_{pq} \hat{n}_p\hat{n}_q + 
    \beta\sum_p h_p \hat{n}_p.
    \label{eq:QUBO_hamiltonian}
\end{equation}
The transverse field encodes the importance of the filtered FMs
\begin{equation}
 h_p=\frac{\tilde{\alpha}_p }{\max \tilde{\alpha}_p},
\end{equation}
such that its values range in $h_p \in [0,1]$. Finally, $\beta$ is a hyperparameter that defines the interplay between the two terms.
For $\beta=1$ all the positive FMs are selected. Conversely, smaller $\beta$ values weaken this bias and encourage QA to explore a broader subset of FMs, thereby incorporating contributions from less dominant gradient components. In the limit $\beta=0$, only cosine similarity is considered, i.e. there are no constraints on the number of selected FMs and the solution will be the trivial null set.

The QA protocol starts with initializing the qubits in the ground state of a driver Hamiltonian  $\hat{H}_{D}=-\sum_{p}^{d}\hat{\sigma}_{x}^{(p)}$, i.e. in the full superposition state of all the bit-strings that encode all possible subsets of FMs $\ket{\psi}_{\mathrm{init}}=\bigotimes_{p=1}^{d} \ket{+}_p$, i.e. $\ket{\psi}_{\text{init}} = \frac{1}{2^{d/2}}\sum_{z\in \{0,1\}^{d}}\ket{z}$.
Here, $\sigma_{x}$ represents the Pauli$-$X operator acting on qubit $p$.
The dynamics is governed by a time-dependent Hamiltonian 
\begin{equation}
    \hat{H}(s)=A(s) \hat{H}_\text{D} + B(s) \hat{H}_{\mathrm{QUBO}},
    \label{eq:HP}
\end{equation}
where $s$ is the dimensionless time parameterized as $s\equiv\frac{t}{\tau}$, where $\tau$ is the total evolution time, $t$ is the time and $A(s)$ and $B(s)$ satisfy the conditions $B(0)=0$, $A(1)=0$. In our case, we chose a linear interpolation between $A$ and $B$, i.e. $A(s)=1-s$ and $B(s)=s$.

Notice that the combinatorial optimization problem grows exponentially with the number of filtered FM $d$ (the Hilbert space has dimensions $2^{d}$), while by solving it with a QA protocol, the final solution after the evolution is sampled using a polynomial number of shots, typically $n_{\text{shots}}=O(d^2)$.

\subsection{Benchmark}

\begin{figure*}[t!]
    \centering
    \includegraphics[width=9cm]{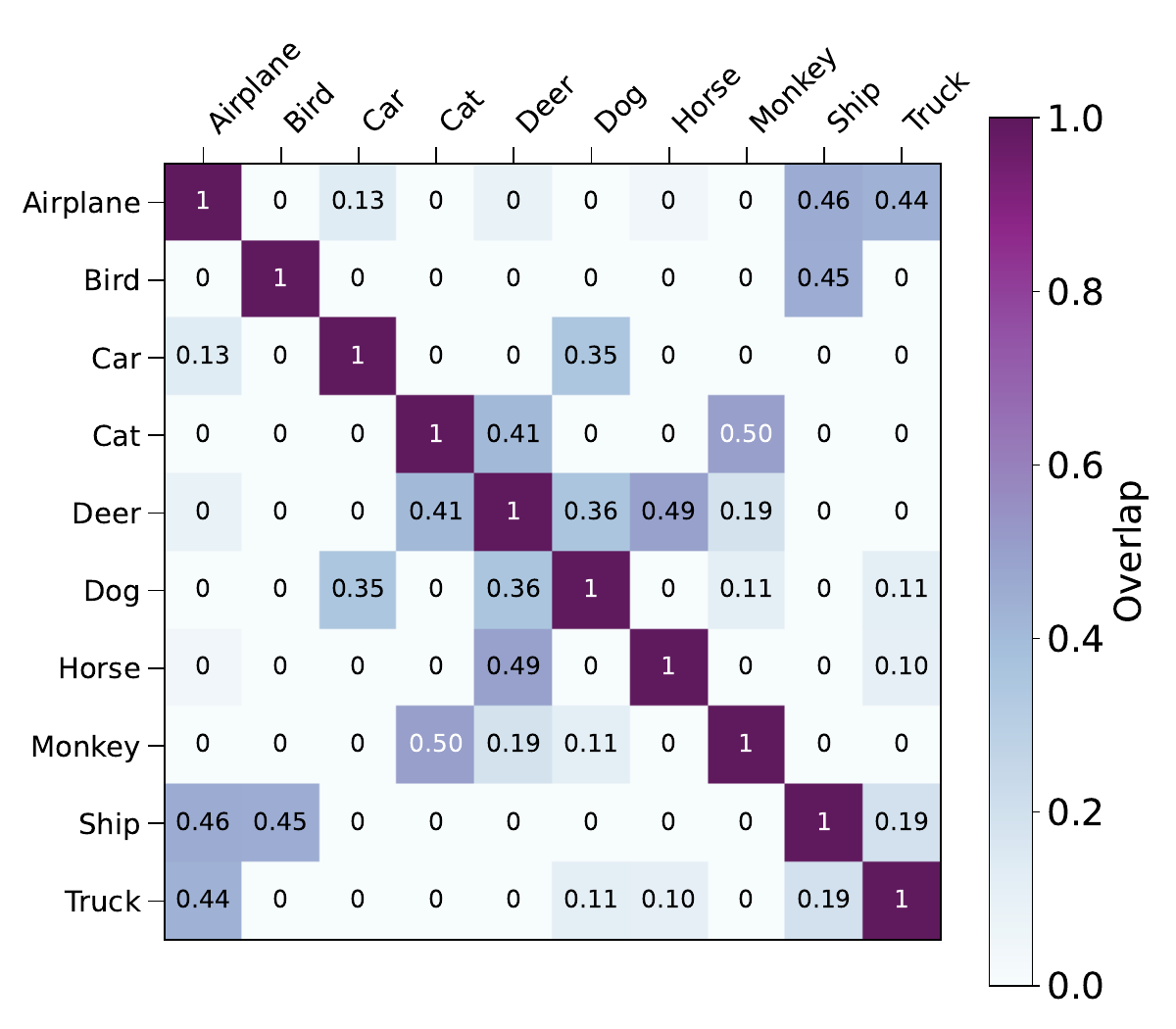}%
    \includegraphics[width=9cm]{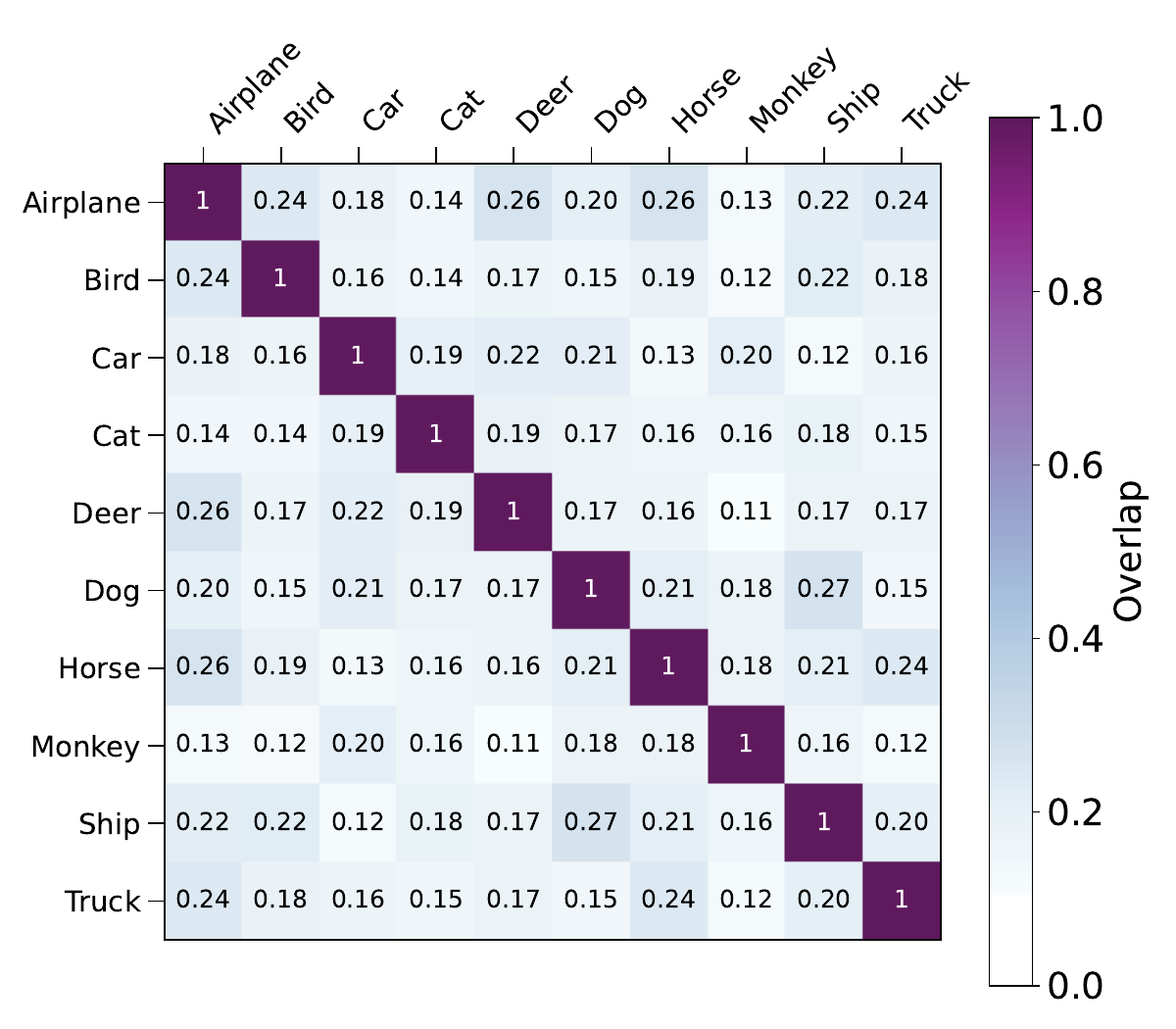}
    \caption{(Left) Class–class correlation map for the FS algorithm for a ResNet-18 constrained to $N_f=16$ FMs in the final layer, with $\beta=0.7$ and trained for 20 epochs.
    Warmer colors indicate stronger overlap between FMs. The matrix is obtained by computing the scalar product between the distributions of the activated FMs per image class, i.e. the Bhattacharya coefficient. For the QA simulation, we set $\tau=50$ and $n_{\text{shots}}=d^2$, where $d$ is the number of FMs that contribute positively to the output's gradient. A full diagonal matrix indicates complete orthogonality among classes. Values below $0.10$ are rounded to $0$.
    (Right) Class–class correlation map for FS algorithm using SA with $\beta=0.99$ for the full ResNet-18 model ($N_f=512$) trained for 20 epochs.}
    \label{fig:CC-corr}
\end{figure*}

\renewcommand{\arraystretch}{1.2} 
\newcolumntype{C}[1]{>{\centering\arraybackslash}p{#1}}
\begin{table*}[t!]
\centering
\begin{tabular}{ %|p{1.6cm}||C{1.8cm}|C{2.4cm}|C{2.7cm}||C{1.8cm}|C{2.4cm}|C{1.9cm}|C{1.9cm}| }
|l||c|c|c||c|c|c|c|}
 \hline
 \multicolumn{8}{|c|}{\textbf{Quantitative analysis - Average Drop $\%$}} \\
 \hline

 \multirow{3}{*}{\textbf{Class}} 
 & \multicolumn{3}{c||}{$\mathbf{N_f = 16}$}
 & \multicolumn{4}{c|}{$\mathbf{N_f = 512}$} \\
 \cline{2-8}

 & \textbf{GradCAM} & \textbf{GradCAM++} & \textbf{QA-$f$GradCAM}
 & \textbf{GradCAM} & \textbf{GradCAM++} & \multicolumn{2}{c|}{\textbf{SA-$f$GradCAM}} \\
 \cline{2-8}
 & - & - & $\mathbf{\beta = 0.7}$ & - & - & $\mathbf{\beta = 0.7}$ & $\mathbf{\beta = 0.99}$ \\
 \hline

 0 Airplane  &13.8  &11.7  &13.8  &1.20  &2.29  &12.9  &2.36 \\ \hline
 1 Bird      &6.66  &3.21  &7.06  &4.76  &4.02  &23.3  &4.01 \\ \hline
 2 Car       &0.47  &0.62  &0.67  &0.97  &1.38  &8.09  &0.85 \\ \hline
 3 Cat       &4.06  &3.83  &4.51  &4.29  &1.29  &14.0  &0.01 \\ \hline
 4 Deer      &12.2  &10.0  &14.6  &5.11  &5.27  &6.47  &4.32 \\ \hline
 5 Dog       &47.7  &36.5 &56.9  &12.1  &17.0  &30.1  &12.8 \\ \hline
 6 Horse     &2.55  &0.87  &3.43  &7.31  &12.4  &32.6  &14.2 \\ \hline
 7 Monkey    &4.60  &1.36  &4.18  &13.0  &18.9  &34.0  &20.4 \\ \hline
 8 Ship      &0.47  &0.12  &0.70  &0.97  &1.59  &14.9  &1.32 \\ \hline
 9 Truck     &0.08  &0.25  &0.14  &2.88  &3.01  &7.37  &2.53 \\ \hline

 $\mathbf{\overline{x} \ \pm \ \sigma_{\overline{x}}}$ 
 &9.25 $\pm$ 5.94 
 &6.95 $\pm$ 5.41 
 &$9.70\pm6.13$ 
 &$5.27\pm4.58$ 
 &6.73 $\pm$ 4.79
 &$18.4\pm8.69$
 &$6.28 \pm 4.25$ \\
 \hline

\end{tabular}
\caption{Comparison between GradCAM, GradCAM++ and our QA-$f$GradCAM / SA-$f$GradCAM protocols in terms of the Average Drop $\%$ per image class. The QA simulation takes $\tau=50$ and $n_{\text{shots}} = {d}^2$, where $d$ is the number of filtered FMs for each image. The SA-$f$GradCAM has been executed for two different $\beta$ values ($\beta=0.7, \ 0.99$) to show the impact in the performance of this meta parameter. The values correspond to the mean of the Average Drop $\%$, while the standard deviation of the mean is calculated by computing the average of the standard deviations first, and then computing the standard deviation of the mean. Our proposed methods achieve comparable results as the state-of-the art GradCAM approaches.}
\label{tab:performance}
\end{table*}

\begin{figure}[h!]
    \centering
    \includegraphics[width=0.8\linewidth]{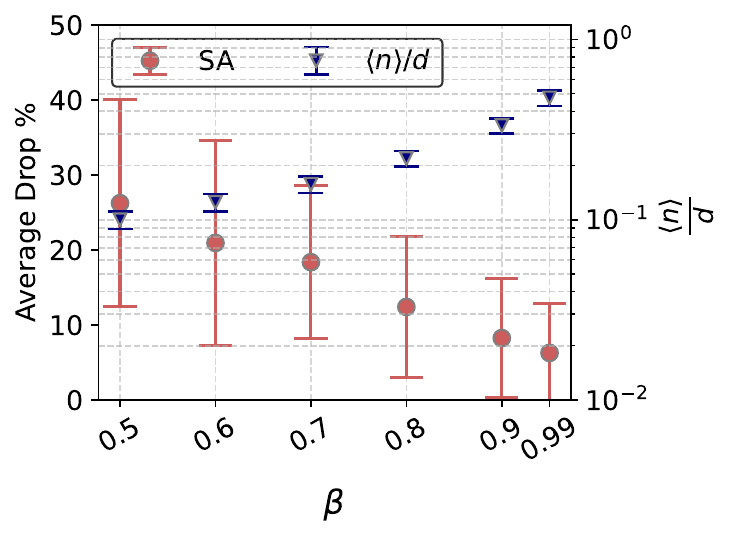}
    \caption{Average Drop \% for SA as a function of $\beta$ (left axis).
    Ratio between the elements within the solution bit-string and the activated qubits per problem Hamiltonian (right axis). By increasing $\beta$, the linear term becomes predominant and the Average Drop $\%$ reduces, while the ratio increases since many more FMs are selected.}
    \label{fig:avgdrop_sa}
\end{figure}

To validate the proposed method, we take the STL-10 \cite{coates2011analysis} dataset that contains 5000 images in total that belong to 10 distinct classes.
As a CNN architecture, we use a pretrained ResNet-18.
To study the scalability of our approach, we benchmark the QA simulations by progressively increasing $N_f$ from 16 up to 32 FMs in the last convolutional layer. This restriction is necessary to keep the exact diagonalization of the QUBO Hamiltonian (required for the physical analysis of the algorithm) computationally feasible, as the Hilbert space grows exponentially with $N_f$. For the full ResNet-18 with $N_f=512$ FMs, where exact diagonalization becomes intractable, we resort to SA to sample the solution bitstring and estimate the protocol's performance at larger scale.

We start by obtaining the confusion matrix by computing the Bhattacharyya coefficient~\cite{patra2015new}, which estimates how likely two classes share the same FMs.
A fully diagonal correlation matrix indicates well-separated (i.e., disentangled) FMs that capture information specific to a single image class. In contrast, non-zero off-diagonal elements reveal shared representations among different image classes and constitute the primary source of false-positive and true-negative errors.

Figure~\ref{fig:CC-corr} shows the class-class correlation map for a CNN restricted to $N_f=16$ filtered FMs using QA (left) and for the full ResNet-18 with all the $N_f=512$ FMs where we adopted SA (right). 
For the QA we set $\beta=0.7$ and $\tau=50$, while for the SA we use $\beta=0.99$, as explained below.
In both cases, the matrix shows an overall good disentanglement across different image classes.
Focusing on the $N_f=16$ case, we obtain only $14$ non-zero ($\geq 0.10$) overlap out of the $45$ possible image class combinations, and all of these lead to a Bhattacharya coefficient below $0.50$.
For the full FMs simulation with SA, we obtain an almost uniform overlap with values that still are below $0.30$ without sparse peaks. This result indicates that we achieve an overall disentanglement with the classes. 

We then test our proposed FS against well-known classical methods that have achieved promising results and constitute the state-of-the-art in the context of interpretability. Specifically, we compare the explanation maps obtained by our protocol with $N_f=16$ against the spatial projections originated by GradCAM and GradCAM++ for the entire set of image classes. 
The explanation map derived by the QA-$f$GradCAM is a bit-string containing the FMs that satisfy the constraints of the QUBO problem (positive gradient contribution and maximal orthogonality between FMs), as explained in the previous section. In this way,  we expect to obtain FMs that are less spurious and redundant compared to GradCAM and GradCAM++ since they do not have this constraining term.

For a quantitative analysis, we estimate the $\text{Average Drop}\  \%$ ~\cite{chattopadhay2018gradcampp}, which essentially replaces all pixels excluded by the explanation map with 0 (i.e. sets them to black), then inputs this masked image to the model to calculate the prediction score. A lower $\text{Average Drop}\  \%$ indicates that the explanation map captures the most relevant regions and that masking the remainder does not significantly affect accuracy. Table~\ref{tab:performance} shows the $\text{Average Drop}\  \%$ per class for the different methods. QA-$f$GradCAM applied to $N_f=16$ features with $\beta=0.7$ exhibits a comparable average value with respect to classical GradCAM and GradCAM++. For the full ResNet-18 with $N_f=512$, SA produces explanation maps with a higher $\text{Average Drop}\  \%$ than GradCAM and GradCAM++ when $\beta = 0.7$. 

We investigate the role of $\beta$ as a potential meta-parameter that can be used to increase the model's accuracy. Fig.~\ref{fig:avgdrop_sa} shows the $\text{Average Drop}\  \%$ and the mean of the ratio between the selected FMs $n$ over $d$, $\frac{\langle n \rangle}{d}$ as a function of $\beta$. As $\beta$ increases, the linear term in the $\hat{H}_{QUBO}$ Hamiltonian dominates and more FMs are selected, reflected in the growing ratio. This underscores the importance of tuning $\beta$ until the $\text{Average Drop}\  \%$ converges to that of the reference CAMs. In our benchmark, $\beta = 0.99$ is the optimal value for the full ResNet-18 $N_f=512$, as shown in the rightest column from Tab.~\ref{tab:performance}. 

\subsection{Model's efficiency and probability of success}

\begin{figure}[t!]
    \centering
    \includegraphics[width=0.8\linewidth]{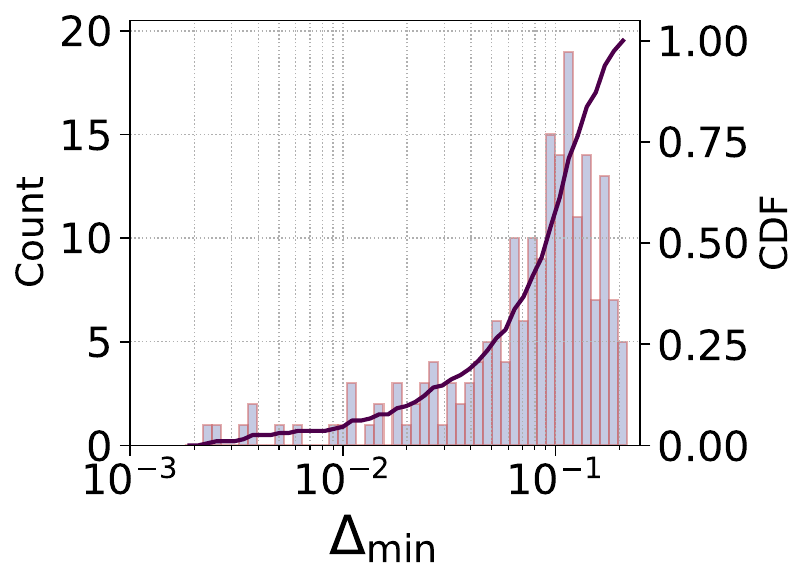}
    \caption{Energy gap $\Delta_{\min}$ distribution for $N_f=16$, for $\tau=50$ and $\Delta s=0.01$ across all 200 samples of the test set. In purple, the cumulative distribution tells us about how likely we find values of the minimum gap.
    }
\label{fig:delta_cummulative}
\end{figure}

\begin{figure}[t!]
    \centering
    \includegraphics[width=0.9\linewidth]{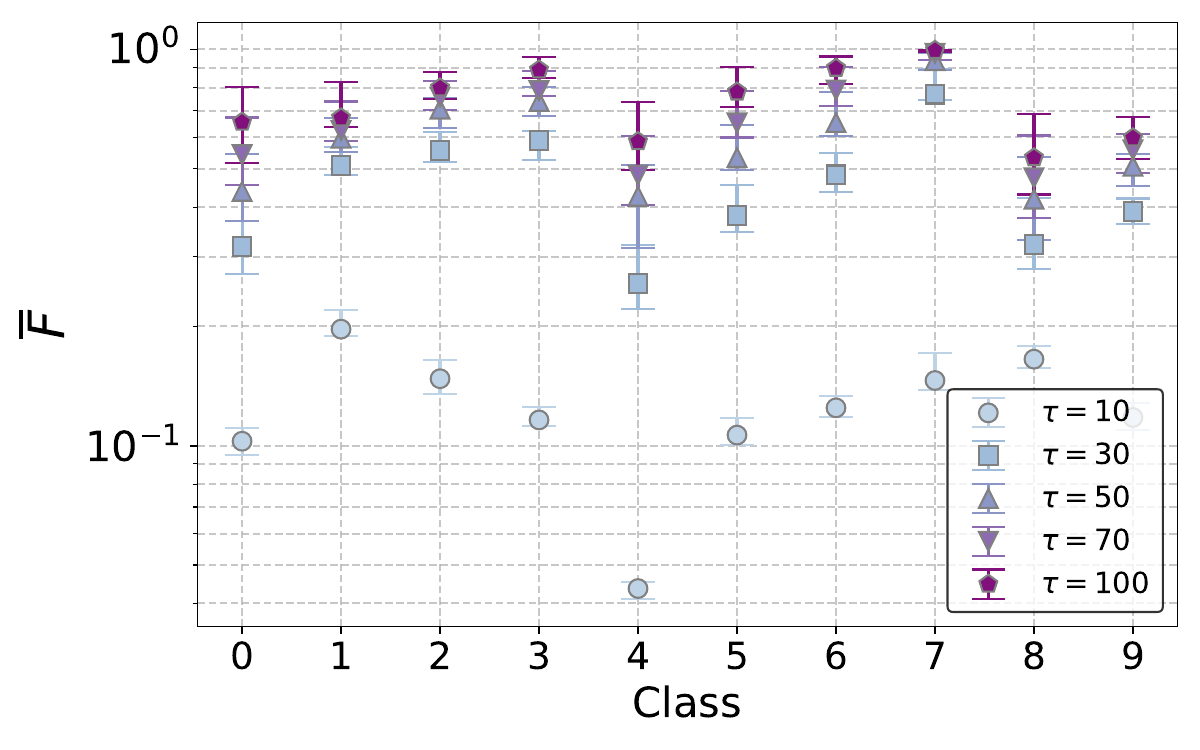}
    \includegraphics[width=0.9\linewidth]{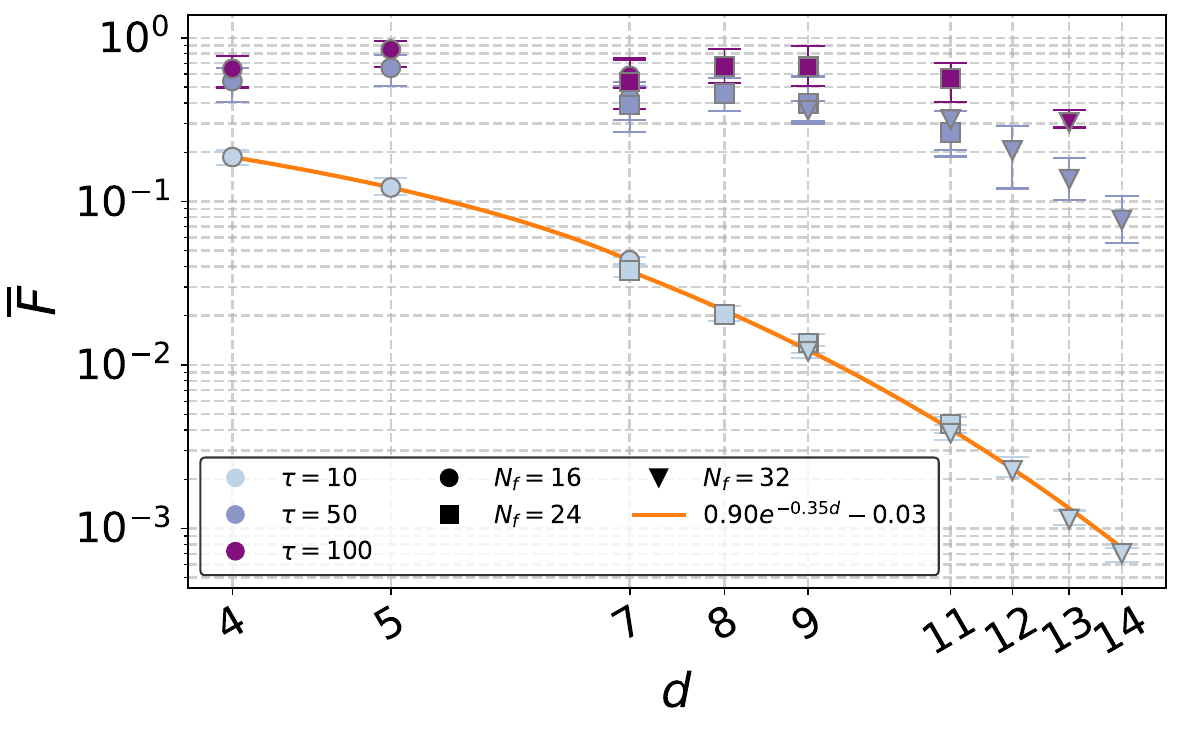}
    \includegraphics[width=1\linewidth]{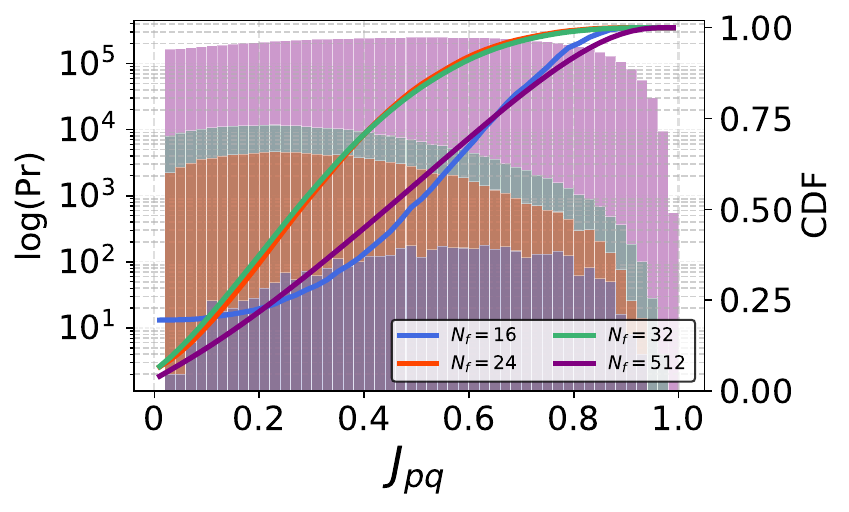}
    \caption{(Top) Median fidelity $\overline{F}$ values as a function of the class for different values of $\tau$ and $N_f=16$. For lower values of $\tau$ it exhibits a scaling behavior approximately proportional to $\sim \frac{1}{d}$, indicating increased difficulty as the problem size grows, as expected. Given that the distribution of the minimum gap is shifted towards larger values, as $\tau$ increases, we notice a suppression of this scaling as the evolution achieves an adiabatic regime.
    (Medium) Median $\overline{F}$ values as function of the filtered FMs $d$ for distinct $\tau$ with different $N_f = 16, \ 24, \ 32$. For higher values of $\tau$, the fidelity shows a global constant trend, while for $\tau=10$, in diabatic regime, we observe an exponentially decreasing trend, in accordance with the Landau-Zener formula.
    (Bottom) $J_{pq}$ distribution across all the classes of the entire dataset for $N_f = 16, \ 24, \ 32$ and $N_f = 512$. The distribution matches a Gaussian one for low $N_{f}$ and approximates a Poissonian for larger number of FMs. This behavior is compatible with the well-known Sherrington-Kirkpatrick model.}
    \label{fig:fidelity-tau}
\end{figure}

We now proceed to analyze in detail the physical properties of the QA protocol to address its efficiency and probability of success. In QA, a relevant quantity to determine the model's complexity is the energy gap $\Delta_{\min}\equiv E_{1}-E_{0}$ between the ground state and 1st excited state energies of the time-dependent Hamiltonian $\hat{H}(s)$ from Eq.~\eqref{eq:HP}. If $\Delta_{\min}$ tends to zero, the annealing time parameterized by $s$ must be longer to ensure that the system remains in its ground state.

We start with an aggregated analysis of the gap for all the dataset. For each image, we diagonalize $\hat{H}(s)$ to obtain its spectrum and compute the $\Delta_{\min}$ for time steps of $\Delta s = 0.01$ and $\tau=10$. In Fig.~\ref{fig:delta_cummulative} we plot the histogram and the cumulative curve.
We observe that the overall trend is placed towards large energy gap values, near $10^{-1}$ units of energy, while the
probability of having values below $10^{-3}$ is very close to $0$. This indicates that the QA evolution is sufficient and robust to converge to the ground state of the target problem Hamiltonian $\hat{H}_\text{QUBO}$ for most of the instances of the QUBO problem.

Next, we investigate the probability of success, i.e. of finding the true ground sate, of the QA protocol for each image. We define the fidelity between pure quantum states (i.e. a measure of distance)
\begin{equation}
    F = \left| \langle \psi_{fin} \mid \psi_0 \rangle \right|^2,
\end{equation}
where $\ket{\psi_0}$ corresponds to the ground state of the problem Hamiltonian $\hat{H}_P$, while $\ket{\psi_{fin}}$ is the quantum state at the end of the annealing evolution, and we use it as a metric that informs us about the success of the QA.
For a random final state, $\bar{F}\simeq \frac{1}{2^{d}}$, represents a uniform distribution. In general, any scaling of $F$ of the type $F \propto d^{-\gamma}$, with $\gamma >0$ is considered a good result, since it restricts the research method to a subspace polynomial with the degrees of freedom of the system.

One would expect that for low values of $\tau$, the evolution will not be adiabatic and the system will end up in an excited state instead of the ground state. Therefore, this measure will also illustrate what is the scale of the annealing time $\tau$ necessary to archive accurate results.
Figure \ref{fig:fidelity-tau} (top) shows the median value $\overline{F}$ for each class of the dataset and for different values of $\tau$. Since the number of filtered FMs $d$ may vary from image to image and that measure is what determines the number of qubits of the QA protocol, we also compute $\overline{F}$ as a function of $d$ and show it in Fig.\ref{fig:fidelity-tau} (bottom).
In both figures, the error bars represent the $1^{st}$ and $3^{rd}$ quartile of the samples.
For $\tau=10$ the system falls into a diabatic regime, i.e. too fast, and the result does not correspond to the ground state. In general, we observe that for $\tau>50$ the solution is obtained with higher probability. Also, there is no significant difference between the behavior as a function of the class or as a function of the number of filtered FMs $d$. Therefore, although the dimension of the Hilbert space increases exponentially for $d$, we still obtain higher fidelity values overall, meaning the QA is able to end up into the ground state of $\hat{H}_{\text{QUBO}}$ for the majority of the problem instances.

Finally, we analyze the structure of the couplings $J_{pq}$. As visible at the bottom part of Fig.~\ref{fig:fidelity-tau}, the couplings distribution is well approximated by a Gaussian, meaning that our Ising model falls into the well-known Sherrington-Kirkpatrick (SK) model \cite{das2005quantum, panchenko2013sherrington}. Before displaying, we removed the bins representing the diagonal values of the cosine similarity. Subsequently, we focus on distinct sizes of our problem to see how the distribution changes as the number of FMs increases. For $N_f = 16$, the cosine similarity is well approximated by a Gaussian distribution, while for larger numbers of FMs, we obtain a distribution similar to a Poissonian.

\section{Discussion}\label{sec:discussion}

DL architectures such as ResNet-18 can produce hundreds of FMs, $N$, at a single convolutional layer. Identifying the most informative and non-redundant subset for a given prediction requires searching through a large solution space that scales $\sim 2^{N}$, and
classical FS algorithms may struggle to scale efficiently under such conditions. This fact motivates the use of quantum computers, in particular QA, machines specially designed to solve combinatorial optimization problems by sampling from the problem solution obtained after the quantum evolution.

Our work introduces a novel approach to assess interpretability of large models, such as CNNs, by characterizing the learned FMs extracted from a target convolutional layer, and efficiently selecting those that contribute to select class-specific information from the data samples.
Our approach is motivated by the desire of performing a characterization of the entire set of FMs that compose the image. Each of them is appropriately inspected, revealing for which image class it is activated and whether it is shared among distinct images. We also achieve a significant model simplification, incrementing the level of confidence in interpreting predictive outcomes, by directly modeling the learned set of representations instead of focusing on the architecture itself.
To do so, we apply a FS algorithm to perform FMs characterization. Subsequently, we reformulate the FS algorithm as an energy minimization problem to be solved by a QA.

As shown in Fig.~\ref{fig:CC-corr}, while obtaining a quite disentangled correlation matrix, our model also learns representations that are commonly found in distinct image classes. For instance, classes representing animals may show a non-zero overlap since many of them are characterized by the same number of legs, similar color patterns or similar overall shape. Similarly, airplane images may be confused as ships or trucks because the overall shape of singular structures present in both classes is similar. 
In addition, we are assuming all the FMs that are selected belong to the subject of the classification. Rather, it may happen that a positive contribution to the gradient is obtained from a FM that is part of the background, and its selection is enforced through the quadratic term (it may be a geometrically independent vector with respect to those that contain information of the image subject). In this case, we can have distinct background-related FMs that are shared and activated within distinct image classes.  
Nevertheless, the feature characterization serves at this scope: understanding which filters are activated among distinct classes to further drive a more sophisticated training process that would take this into account.

To quantitative benchmark the interpretability of the selected FMs, we evaluate the Average Drop $\%$ of the resulting CAMs. As shown in Tab.~\ref{tab:performance}, the quantum FS with $N_f=16$ features achieves a comparable Average Drop as full GradCAM and GradCAM++ and thus to the interplay parameter $\beta$ in the QUBO formulation from Eq.~\eqref{eq:QUBO_hamiltonian}, as it is observed in the results obtained in the $N_f=512$ model, also presented in Tab.~\ref{tab:performance}. Nevertheless, the Average Drop $\%$ metric is sensitive to the number of selected FMs and only measures the variation in model confidence when the regions considered most relevant are preserved while the remaining image content is masked. Consequently, a lower Average Drop $\%$ does not necessarily correspond to a more interpretable or faithful explanation.

In our framework, the QUBO reformulation of the FS problem operates directly on the hidden representations learned by the model during training, selecting only those FMs that satisfy the imposed optimization constraints. Unlike standard Grad-CAM approaches, the selected FMs are not necessarily aggregated, allowing each representation to be analyzed individually. Moreover, the quadratic term enforcing mutual geometric orthogonality among FMs encourages the selection of diverse and non-redundant representations, which may also include information not strictly related to the main image subject. However, this formulation enables the disentanglement of information learned at different stages of the network, ultimately facilitating a more explicit characterization of the selected features.

For the targeted convolutional block of the full ResNet-18, one needs to fine-tune the $\beta$ parameters from the QUBO Hamiltonian to obtain comparable results as in GradCAM and GradCAM++. Two compounding factors explain this fact. First, the spatial resolution of the FMs extracted from the last convolutional block of the full ResNet-18 is lower than in the custom architecture ($N_f=16$), leading to larger pairwise cosine similarities $J_{pq}$ and a more aggressive selection for the same $\beta$, so that the quadratic redundancy term in $\hat{H}_\mathrm{QUBO}$ suppresses a larger fraction of positively weighted FMs, leaving only a small surviving subset. Second, at $N_f=512$ the QUBO problem is too large to verify whether SA reaches the global minimum, since neither simulated QA nor brute-force search scale to this problem size, so suboptimal solutions may also contribute to the degraded performance. As discussed in the Results, increasing $\beta$ toward 0.99 partially compensates for this by making the gradient importance term dominant, though the exact GradCAM reference value is not fully recovered since the orthogonality constraint on the selected FMs persists with reduced weight. Introducing a cardinality penalty constraining the number of selected FMs while leaving the redundancy term unaffected could offer a more principled alternative, improving the explanation maps without sacrificing the diversity of selected representations.

Next, we assessed the complexity of the QA protocol. The minimum energy gap $\Delta_\mathrm{min}$ across all test images is concentrated near $10^{-1}$ energy units, with essentially zero probability of finding $\Delta_\mathrm{min} < 10^{-3}$, indicating that the QA evolution robustly converges to the ground state of $\hat{H}_\mathrm{QUBO}$ for the vast majority of problem instances. The fidelity between the final annealed state and the true ground state remains high and approximately constant across all image classes and problem sizes for sufficiently large annealing times $\tau$, while for $\tau<10$ the system falls into the diabatic regime and the fidelity decays with the number of filtered FMs $d$. The average $\Delta_\mathrm{min}$ scales as $O(d^{-1})$, implying that the required annealing time scales only quadratically with $d$, a favorable polynomial scaling, in contrast to the exponential worst-case complexity known for general QUBO problems~\cite{Barahona_1982}. This suggests that the particular structure of the QUBO instances arising from FM selection makes the problem tractable for QA in practice. 

In terms of model's complexity, we also analyze the couplings $J_{pq}$ (Fig.~\ref{fig:fidelity-tau} bottom). The fact that the distribution is unstructured and represents a Gaussian means that our Ising model from the QUBO formulation falls into SK model \cite{das2005quantum, panchenko2013sherrington}, which, in its hardest instances, is NP-Hard \cite{Barahona_1982}. 
This result is consistent with the well-known averaging effect in deep networks: as the number of FMs grows, individual feature representations become increasingly correlated and their pairwise overlaps self-average, suppressing large fluctuations. As a result, the effective QUBO instance obtained in our benchmark yields a tractable optimization landscape. Tuning $\beta$ parameter may increase the complexity of the problem if combined with a more structured $J_{pq}$ distribution. In fact, other datasets can present more sophisticated coupling distributions that could not be estimated in a classical way. 

Applying the proposed methodology to earlier convolutional blocks, where FMs do not yet exhibit a clearly interpretable semantic structure, could improve the activation of those FMs that significantly contribute to the final prediction. 
Performing FS at these early stages could effectively act as a structured information filtering mechanism by suppressing redundant FMs before higher-level representations are formed so the CNN could be encouraged to propagate only the most relevant low-level features. 

Although our approach is model-dependent, meaning results can vary with different model structures or training setups, it is flexible and can be extended to a wide range of alternative unsupervised and generative learning models. As explained above, our work is universal: the FMs selection procedure can be sequentially applied to any intermediate convolutional block to study which trainable filters are activated across the whole network. We can also use our approach to reverse-engineer a FM search by penalizing the selection of orthogonal FMs and thus restricting our model to find those that are maximally aggregated. In conclusion, our novel approach to address the FS problem in DL and the QA mechanism that can deal with large optimization landscapes is flexible enough to be extended as an agnostic tool for explainable AI.

\section{Methods}\label{sec:methods}

\subsection{Dataset}\label{sec:dataset}

We use the STL-10 dataset \cite{coates2011analysis}, which is a dataset used primarily for image classification. It consists of $10$ distinct classes (airplane, bird, car, cat, deer, dog, horse, monkey, ship, truck) of $96\times {96}$ pixels RGB images, with $4750$ training and $250$ test samples. We extract $250$ images from the training set to construct the validation set. All the images are acquired from labeled examples from ImageNet. We upsample the original images to a dimension of $224 \times 224$, since this corresponds to the dimension of the data samples within ImageNet that are used to train the ResNet-18.

\subsection{CNN architecture}\label{sec:cnn-architecture}

As CNN architecture, we use a pretrained ResNet-18 \cite{he2016deep}, originally trained on the ImageNet dataset \cite{deng2009imagenet}. For transfer learning, we replace the last convolutional block and the classification tail with custom modules, ensuring a reduced number of final class-specific features, i.e. from $256$ to $N_f$. 
Specifically, we build 
such layer preserving the structure of the original ResNet layer, including BatchNorm and ReLU activations after two \textit{Conv} operations that take as input 256 FMs and output 256 in the first, and $N_f$ in the second.
We also introduce ReLU activation functions at the end of each layer. While our algorithm is compatible with the full ResNet-18, reducing the number of final FMs makes the analysis of the QUBO Hamiltonian computationally feasible. In the first attempt, we consider the model with a final convolutional block producing $N_f = 16$ FMs. Subsequent scalability experiments involve models with $N_f = 24, \ 32$ FMs. 

For the SA simulation we consider the full ResNet-18 architecture with all FMs ($N_f = 512$). This architecture includes a Global Average Pooling layer at the end that compresses the entire information to $N_f = 512*1*1$.

\subsection{Deep Learning setup} \label{sec:dlsetup}

Transfer learning is performed using the PyTorch library \cite{paszke2019pytorch}. We adopt the Adam optimizer \cite{kingma2015adam} to fine-tune the parameters of the additional convolutional block and the classification tail. The convolutional block consists of a sequence of $256$ input channels and $N_f=16$ (in the scalability experiments, we increase it up to 32) output channels, followed by batch normalization and a ReLU activation function. The classification tail is a dense layer of dimension
$H_{f}\times W_{f}\times N_{f}$, where $H_{f}\times W_{f}$ is the spatial size of the FMs.

\begin{figure}
    \centering
    \includegraphics[width=1\linewidth]{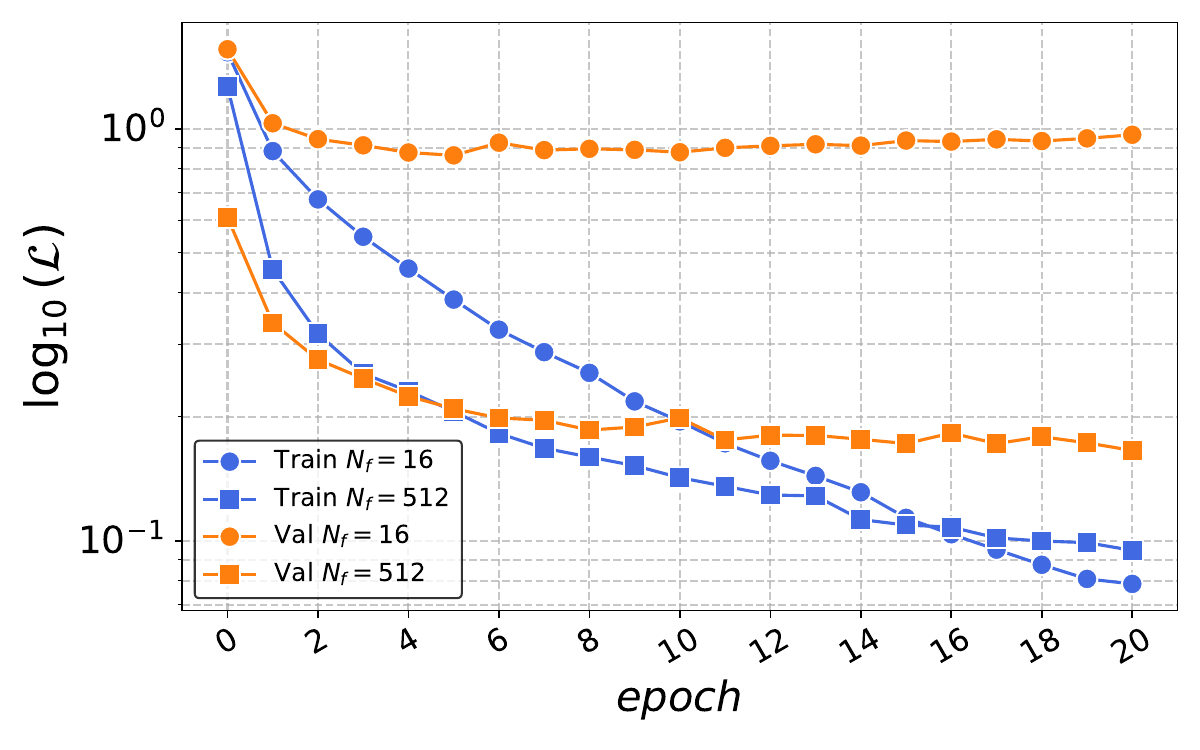}
    \caption{Model training checkpoints. The train and validation losses of the model's tail are computed at each epoch from a total of $N_t = 20$. The plot depicts the logarithm of the training and validation (cross-entropy) loss for two distinct models composed of $N_f = 16$ and $N_f = 512$ FMs in the last convolutional block respectively.}
    \label{fig:checkpoint}
\end{figure}

We trained the network for $N_t=20$ epochs and we store the value of the validation accuracy after each epoch (see Fig.~\ref{fig:checkpoint}). We then consider the model configuration that has reached the highest value of validation accuracy to avoid overfitting, and we finally perform the evaluation on the test set, obtaining a test accuracy of $acc = 94\%$.

After model training, the most important part at this stage is the computation of the output gradient with respect to the single FM extracted by the convolutional wedge, which is done by the PyTorch built-in \textit{hook} method. It is a powerful mechanism for gaining insights into the behavior of the CNN during both forward and backward passes. We use the \textit{forward hook} that allows us to retain the output of the model after a single forward pass. By hooking the model, we are able to compute the gradient of the model's output with respect to the activation. After the modified gradient computation, we eliminate the corresponding hook to move to a subsequent data point.

The last important detail to consider is that we use two different versions of the same ResNet-18 model, as mentioned in \ref{sec:cnn-architecture}. In the first version we consider the FMs extracted from the custom convolutional layer inserted after the third layer of the original ResNet-18, while for the second version we get the full set of FMs from the fourth original layer, before the classifier. In the first version, the FMs belong to the third layer and the information is compressed to originate a number $N_f$ of these ones. Thus, their resolution is different with respect to those that belong to the original full ResNet-18.

\subsection{Average drop $\%$}\label{sec:average_drop}

In our benchmark analysis we compute the $\text{Average Drop} \ \%$~\cite{chattopadhay2018gradcampp}, which calculates the percentage of drop in the model's confidence for a particular image class when is represented by its explanation map. It is formally defined as 
\begin{equation}
    \text{Average Drop} \ \% = \frac{1}{M} \sum_i^M\frac{\max(0, Y^c(\mathbf{x}_i) - E^c(\mathbf{x}_i))}{Y^c(\mathbf{x}_i)}\cdot 100,
\end{equation}
where, $c$ corresponds to a particular class, $Y^c(\mathbf{x}_i)$ refers to the model's output score (confidence) on the image $\mathbf{x}_i$, and $E^c(\mathbf{x}_i)$ is the same model's confidence with only the explanation map region as input. If the Average Drop computed on the selected FM subset exceeds that of the full GradCAM, the subset fails to capture the discriminative information carried by the complete set. Conversely, a lower Average Drop indicates that removing redundant FMs improves representability, validating the selection.

\subsection{Quantum annealing simulation}\label{sec:QA}

The QA simulation is done using matrix and tensor product multiplication to build the global time-dependent Hamiltoninan $\hat{H}(s)$ from Eq.~\eqref{eq:HP}. From the $\hat{n}$ operators, the matrix exponentiation built-in method from the \textit{Scipy} \cite{2020SciPy-NMeth} library is used to perform the quantum state evolution in the same fashion of the Suzuki-Trotter formula.
To analyze $\Delta_{\text{min}}$ and compute the fidelity $F$, we apply the Lanczos diagonalization method \cite{dehesa1981lanczos} to extract the eigenvalues and eigenstates of $\hat{H}(s)$ at each time step $s = \frac{\tau}{0.01}$.

\subsection{Simulated annealing}

We build a modified version of the full ResNet-18 model where we attach ReLU activation layers after each convolutional block, as described in \ref{sec:cnn-architecture}. We then extract the entire set of FMs composed of $512$ elements and we apply the QUBO formulation to obtain the bitstring corresponding to the problem solution. The linear term of the QUBO matrix is composed with the positive gradient contribution of the $d$ FMs, while the off-diagonal terms contain the relative cosine similarity between them. We set a number of \textit{num\_reads}$\ =1000$, \textit{num\_sweeps}$\ =1000$, \textit{num\_sweeps\_per\_beta}$\ =1$, \textit{beta\_schedule\_type}$\ =\textit{linear}$ as default, and each read is generated by one run of the SA algorithm, as reported in the documentation of \textit{D-Wave Ocean}, which is used to build and execute SA.

\subsection{Complexity of the QA protocol}
\label{sec:complexity}

Following the approximate adiabatic theorem~\cite{rajak2023quantum, vznidarivc2005scaling}, the time evolution $\tau$ needed to end in the ground state at the end of the protocol, scales as $\tau \propto \Delta_{\min}^{-2}$. Therefore, one needs to prove that $\Delta_{\min}$ does not decrease exponentially with the system size to conclude that QA can be used to efficiently solve a QUBO problem of the form of Eq.~\eqref{eq:QUBO_hamiltonian}. Otherwise, one would need exponentially long $\tau$ to remain in the ground state.

Regarding the analysis of the values of $\tau$ necessary to achieve good fidelity, we notice in Fig.~\ref{fig:fidelity-tau} that for lower annealing time values $\tau \sim 10$, the model falls into the diabatic regime.
In such condition, the probability of success at the end of the annealing process follows a negative exponential scaling as depicted in the red dashed line. This is in agreement with the Landau-Zener formula \cite{arceci2017dissipative}. In fact, in diabatic regime the probability of passing through an avoided crossing is given by
\begin{equation}
    F = 1 - \text{Prob}_{LZ} = 1 - e^{-\lambda\Delta^2},
\label{landauZiener}
\end{equation}
where $\lambda$ is some constant.
As the energy gap $\Delta_{\min}$ reduces, we expect a maximum $\text{Prob}_{LZ}$ value, meaning that the system likely jumps to higher energy levels rather than following the adiabatic theorem, while for larger $\tau$ values, we remain in the adiabatic condition.

\section*{Code and Data availability}\label{sec:code availability}

Code link \url{https://github.com/checc1/FS_QA}.\\

\section*{Acknowledgements}

E. C. acknowledges funding from the Spanish Ministry for Digital Transformation and the Civil Service of the Spanish Government through the QUANTUM ENIA project call - Quantum Spain, EU, through the Recovery, Transformation and Resilience Plan – NextGenerationEU, within the framework of Digital Spain 2026. A.C.-L. acknowledges funding by the European Union, supported by the EuroHPC Joint Undertaking and its members under the Grant Agreement Nº 101159808, including top-up funding by Ministry for Digital Transformation and the Civil Service of the Spanish Government, and from the grant RYC2022-037769-I funded by MICIU/ AEI/ 10.13039/ 501100011033 and by “ESF+". F.A.V. and M.A.G.B. acknowledge funding from the Maria de Maeztu Units of Excellence Programme CEX2021-001195-M, funded by MICIU/AEI/10.13039/501100011033. B. J-D acknolwedges funding from PID2023-147475NB-I00 funded by MICIU/AEI/10.13039/501100011033 and FEDER, UE, by
grants 2021SGR01095 from Generalitat de Catalunya,
and by Project CEX2024-001451-M of ICCUB (Unidad
de Excelencia María de Maeztu).

\begin{appendix}

\section{More analysis on the complexity in the Quantum Annealing protocol}
\label{QA}

\begin{figure}[h!]
    \centering
    \includegraphics[width=\linewidth]{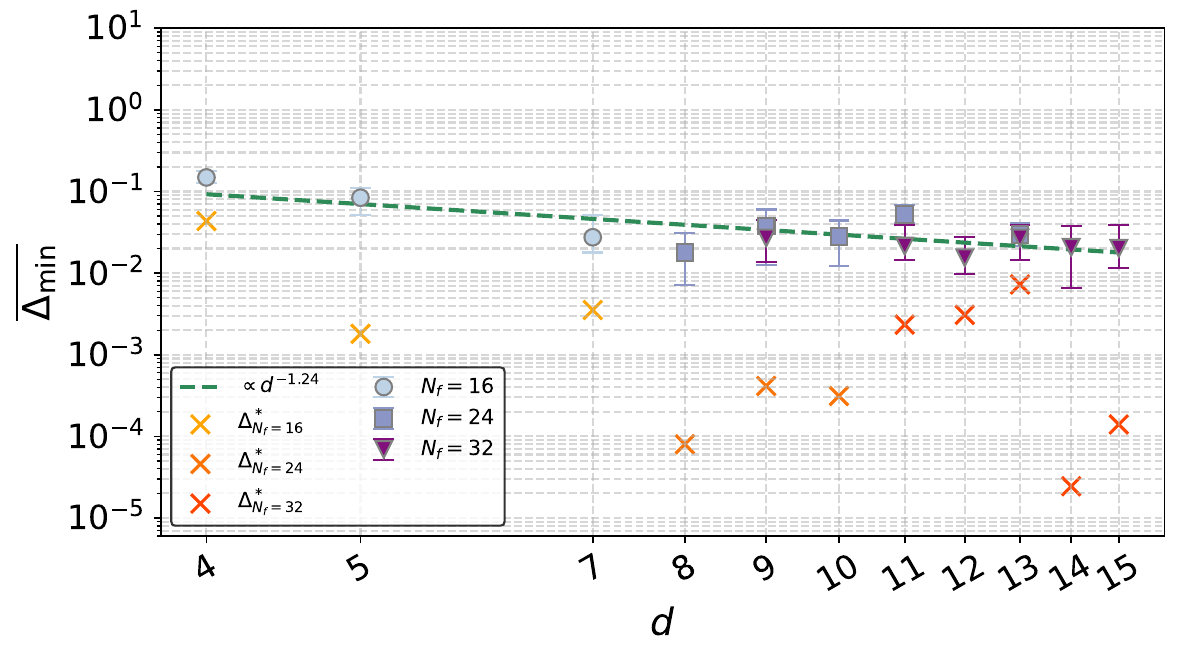}
    \caption{Average of $\Delta_{\text{min}}$ for different $N_f$ as a function of the filtered FMs $d$. The scaling of the gap follows $\Delta_{\text{min}} \propto d^{-1} $ which is estimated by a linear regression. The error bars correspond to the first and third percentile of the entire data values on the specific dimension $d$. The distinct crosses represent the smallest gap $\Delta^*$, i.e. the hardest instances to simulate.}
    \label{fig:gap-scaling}
\end{figure}

The complexity of a QA protocol depends on the energy gap $\Delta_{\text{min}}$. 
We explore the scaling properties of $\Delta_{\text{min}}$ to verify if the QA protocol is feasible on machines with limited $\tau$ and to quantify the advantage of the method. As shown in Fig.~\ref{fig:gap-scaling}, the average $\Delta_{\min}$ for different filtered FMs $d$ scales $O(d^{-1})$, which proves that the scaling of $\tau$ is quadratic with the system size. We also plot the hardest instances, $\Delta^*$, i.e. the cases with the smallest energy gap to show how the it distributes across distinct dimensions $d$. 

\begin{figure}[h!]
    \centering
    \includegraphics[width=\linewidth]{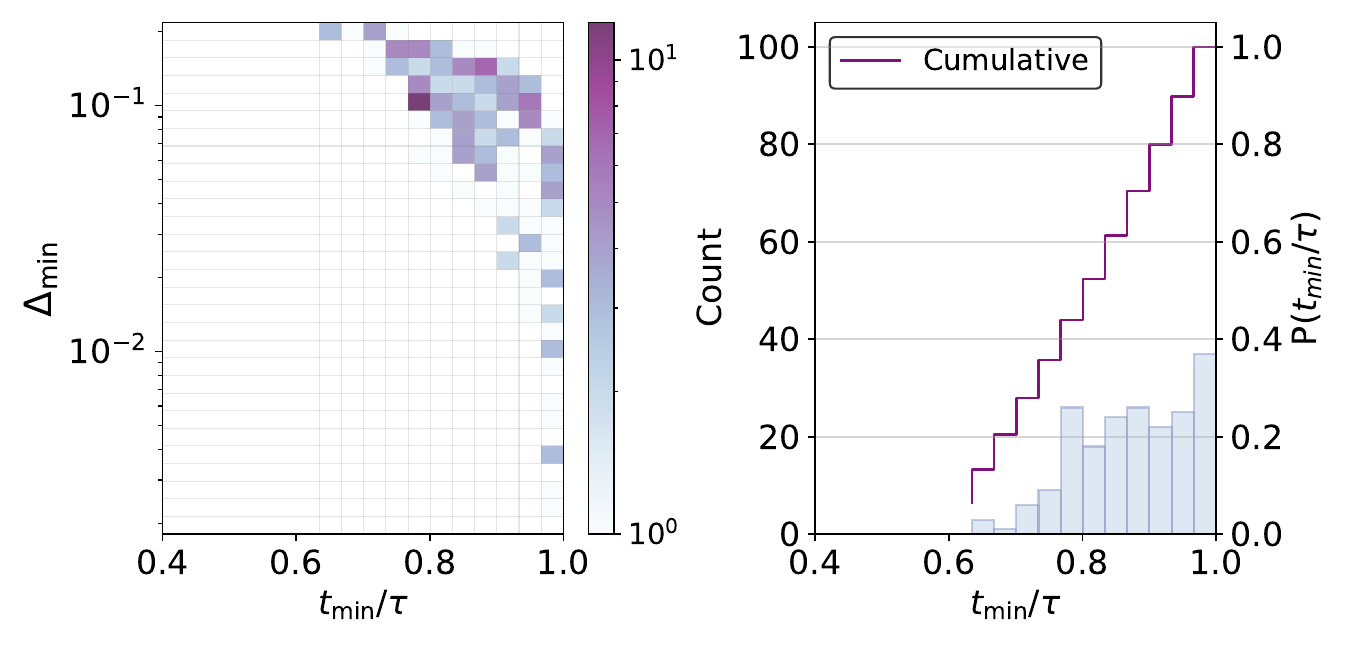}
    \caption{ (Left) Plot of $\Delta_{\text{min}}$ as a function of $t_\text{min} / \tau$. For each problem instance, we run the annealing simulation and we identify the fraction of time where the minimum gap $\Delta_{\text{min}}$ occurs. (Right) Distribution of $t_\text{min} / \tau$ for all the problem instances. From these results, it can be concluded that the minimum gap is found at the end of the annealing process.
    }
    \label{fig:times}
\end{figure}

Another important aspect is to identify the time $t_{\text{min}}\equiv t(\Delta_\text{min})/\tau$ when the evolution finds the minimum energy gap. This information will help in identifying more efficient annealing protocols, i.e. in defining the $A(s)$ and $B(s)$ functions from the $\hat{H}(s)$ Hamiltonian. In Fig.~\ref{fig:times} we plot the distribution of $t_{\text{min}}/{\tau}$ as a function of $\Delta_{\text{min}}$ for all the problem instances of different image classes. We also depict the distribution of the values of time $t_{\text{min}}$, indicating that for the linear scaling of the two Hamiltonian terms, the minimum gap is obtained at the end of the annealing process.

\section{Comment on qualitative observations}

The results shown in the class-class correlation map from the main article reflect certain overlap between different classes, specially for the $N_{f}=16$ benchmark. This section has the goal of illustrating a qualitative analysis about the FMs that are commonly activated and that belong to different image classes. 

\begin{figure*}[t!]
    \centering
    \subfloat[Comparison of the explanation maps between four random samples belonging to Monkey and Cat image class.
    \label{fig:gatti-scimmie-common}]
    {
        \includegraphics[width=\linewidth]{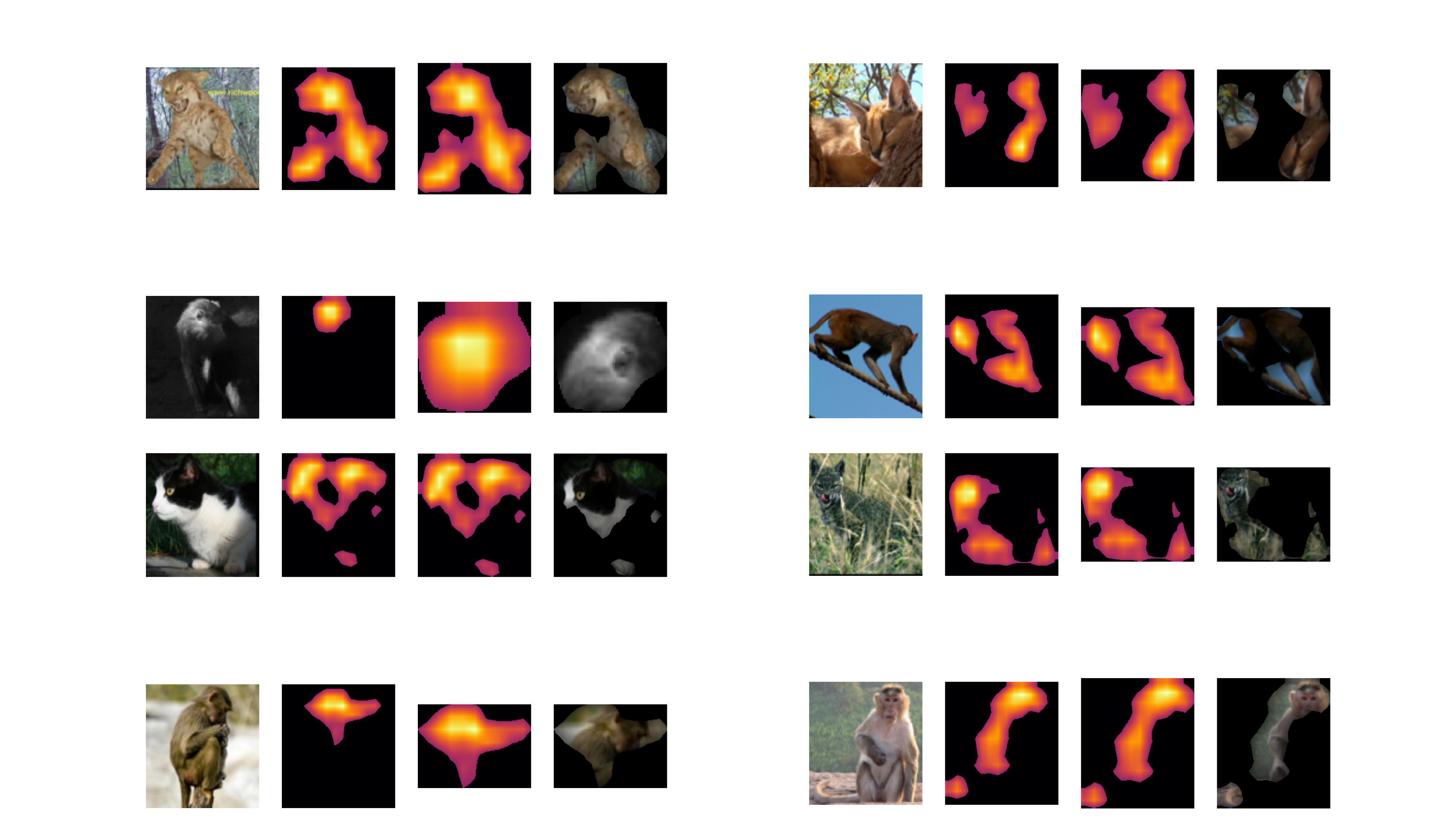}
    }
    \subfloat[Comparison of the explanation maps between four random samples belonging to Deer and Horse image class.
    \label{fig:cavalli-cerbiatti-common}]
    {
        \includegraphics[width=\linewidth]{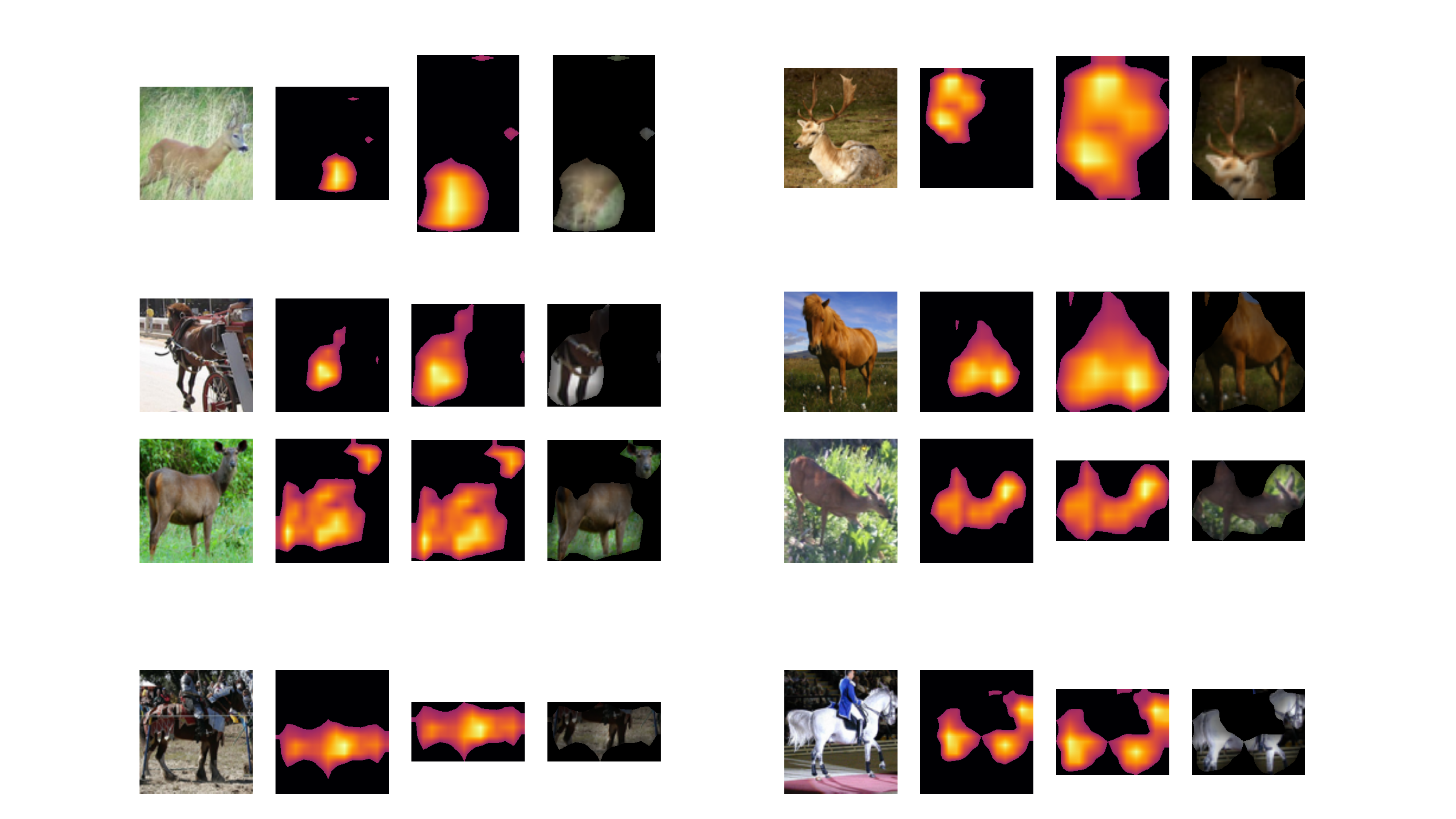}
    }
    \caption{Examples of the explanation maps obtained by considering the FMs shared across distinct image class pairs. Each row shows the original input image, the explanation map colored by gradient importance, its cropped version and the intensity mask on the original image. Each couple of rows are compared.}
    \label{fig:commonFeature}
\end{figure*}

We consider a few samples belonging to different couples of image-classes: respectively Cat (3) - Monkey (7) and Deer (4) - Horse (6), since they show high values of the overlap coefficient.
We then create the corresponding class activation map (CAM) by selecting the FMs within the bit-string solution associated to the samples. At this stage we perform an intersection between the vectors (of different lengths) containing the FMs that are activated for the two classes. In this way we store only those that belong to the bit-string relative to the image we consider.
By running on the pixel values of the relative CAM, we cast those that are below a certain threshold which is set to $0.5$. In this way we obtain the figures displayed in the second column in Fig.~\ref{fig:commonFeature}.
We crop the extracted image by simply restricting the bounding box around the CAM by setting an additional threshold which, in this case, corresponds to $0.6$, but can be increased or lower as wish.

We repeat the same procedure for the two class-items and we crop both the images by using the lowest dimensions of the patch that we extract. The first half of the Fig \ref{fig:commonFeature} displays the common FM shared across Cat and Monkey image classes. In particular, each sub-figure should be read by comparing an entire row of four figures with the one below it. Despite the FM is the same for the image classes we compare, the highlighted region that it corresponds has a different shape, depending on the model's ability of learning patters in distinct image classes. One of the important aspects that we can notice is that the model activates FMs around the main shape of the subjects and the face, except for some samples. In fact, the sub-figure placed at the top-right corner, we observe that the model catches common features that localize on the branches of the tree in the background of the picture portraying the cat, and the legs of the monkey. It seems that for these specific samples the model learns the contrast between the image and the elongated structures of both images, even though they refer to distinct objects.
The sub-figure immediately below instead shows that the model focuses on he entire body-shape of the two animals, and correctly identifies both the subjects.

For the second half of the figure, we compare Deer and Horse. Intuitively these animals are very similar in terms of the main body-shape. Also, the images we have trained the model portray the two animals in similar position in the space. As visible from the first sub-figures, the main filters that are activated represent their legs. Another interesting exception we can observe regard again the sub-figure placed in the right corner of the second half of the main image. The features representing legs and the central body of the brow horse are captured and correspond to the part of the head and the deer's antlers. Two possible interpretations may explain this behavior. First, the model may associate the legs of the horse with structural patterns similar to the deer’s antlers. Second, the main body of the horse may be matched with the brownish background region located behind the deer.

\begin{figure*}[!t]
{\includegraphics[width = 5.5in]{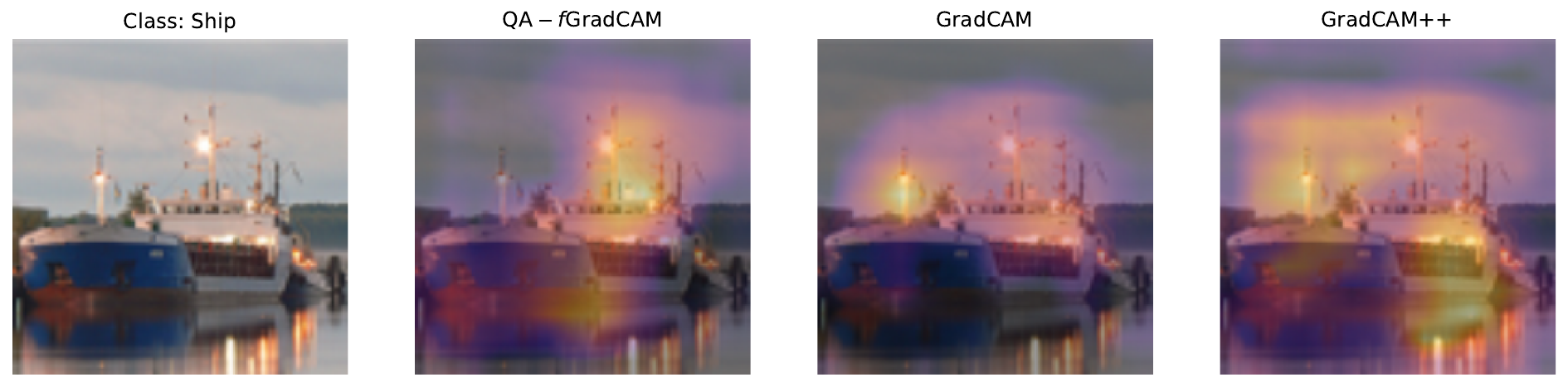}}
{\includegraphics[width = 5.5in]{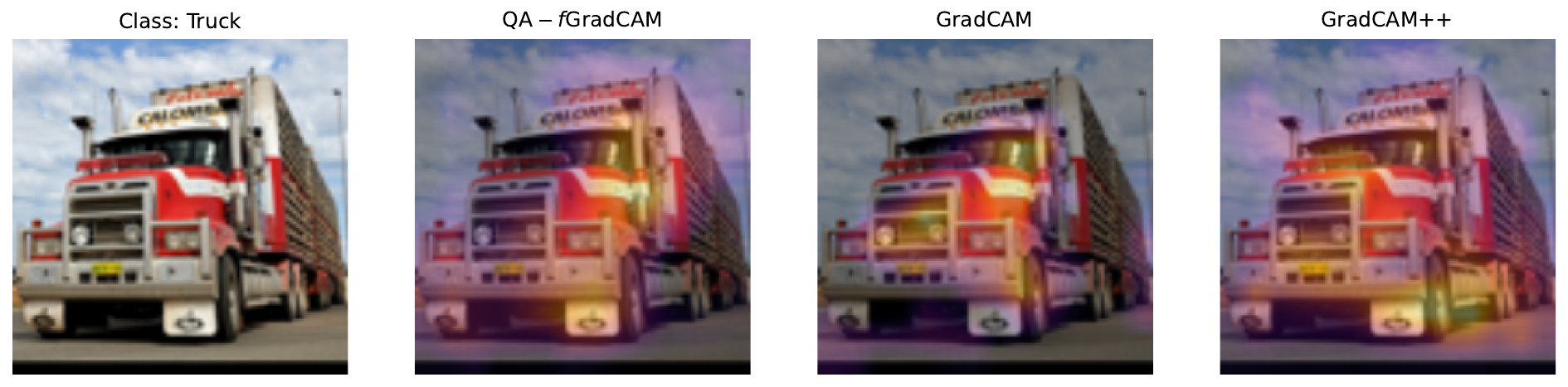}}\\{\includegraphics[width = 5.5in]{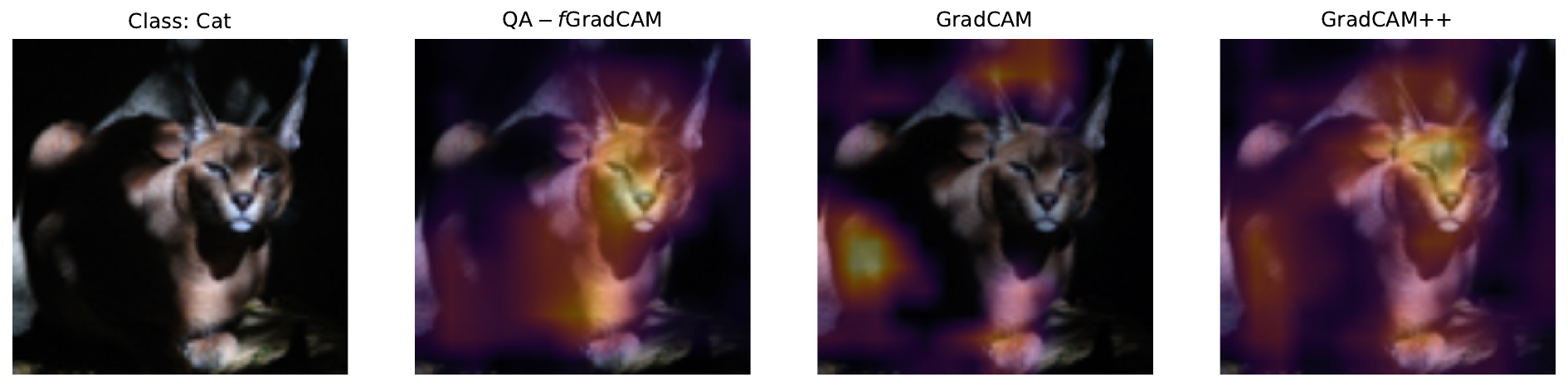}}\\{\includegraphics[width = 5.5in]{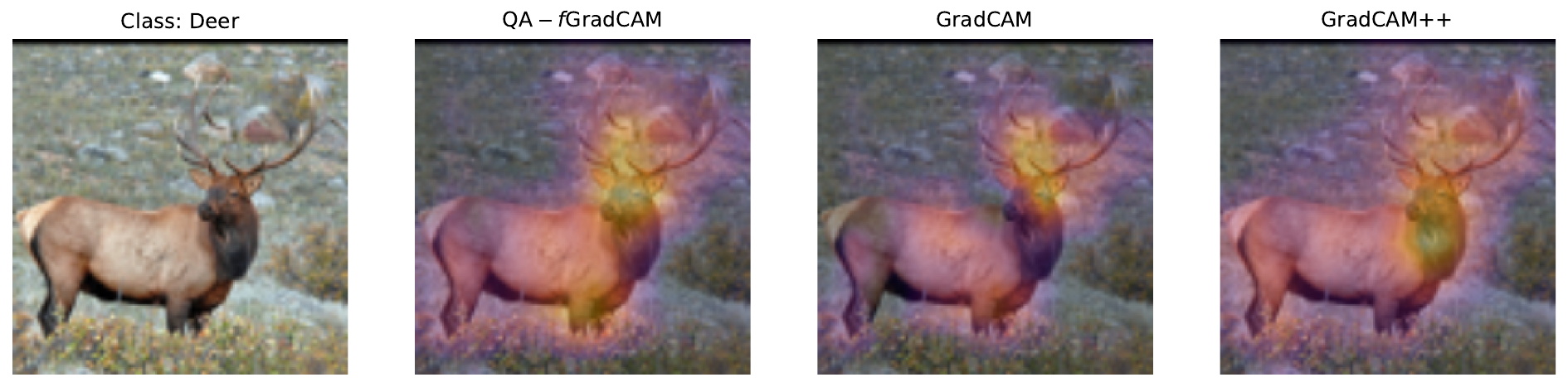}}
{\includegraphics[width = 5.5in]{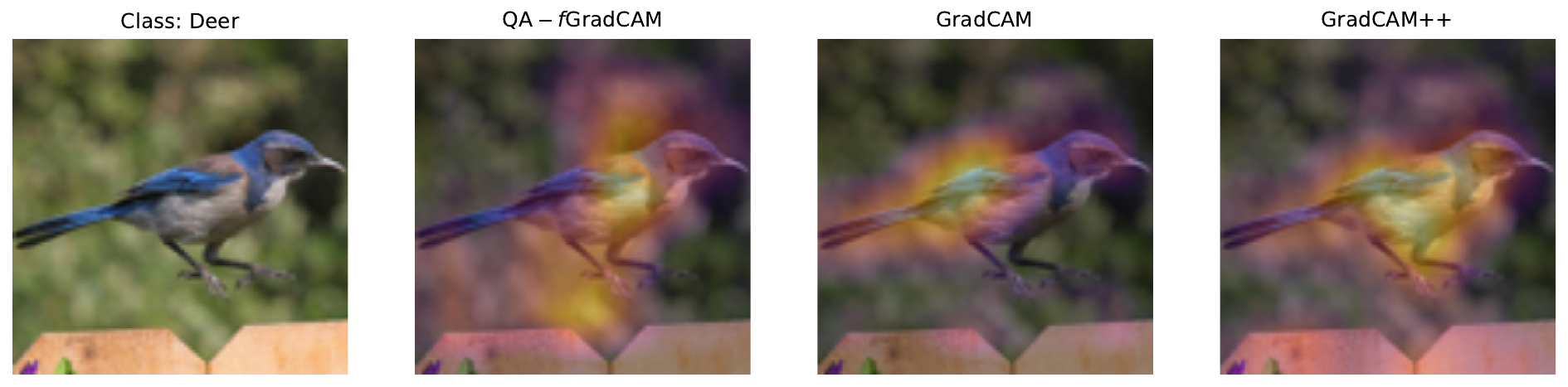}}

\caption{Comparison of the explanation maps obtained with state-of-the-art methods and the QA-$f$GradCAM for four random samples of image class Ship, Truck, Cat, Deer and Bird respectively. }
\label{some example}
\end{figure*}

\end{appendix}

\bibliographystyle{naturemag}
\bibliography{refs}

\end{document}